%% file: main.tex
\documentclass[runningheads]{llncs}

\usepackage{eccv}

\usepackage{eccvabbrv}

\usepackage{graphicx}
\usepackage{booktabs}
\usepackage{wrapfig}

\usepackage[accsupp]{axessibility}  

\usepackage{hyperref}

\usepackage{orcidlink}

\newcommand{\methodname}{SemMorph3D\xspace}
\usepackage{pifont}

\begin{document}

\title{SemMorph3D: Unsupervised Semantic-Aware 3D Morphing via Mesh-Guided Gaussians} 

\titlerunning{SemMorph3D}



\author{Mengtian~Li\inst{1,4} \and
Yunshu~Bai\inst{1} \and
Yimin~Chu\inst{1} \and
Xinru~Guo\inst{1} \and
Haolin~Liu\inst{3} \and
Zhifeng~Xie\inst{1,4} \and
Chaofeng~Chen\inst{2}\textsuperscript{$\dagger$}} 

\authorrunning{M.~Li et al.}

\institute{Shanghai University \and
School of Artificial Intelligence , Wuhan University \and
Tencent \and
Shanghai Engineering Research Center of Motion Picture Special Effects\\}



\maketitle
{\renewcommand{\thefootnote}{}
\footnotetext{\textsuperscript{$\dagger$} Corresponding author.}}
\addtocounter{footnote}{-1}

\input{sec/0_abstract}    
\input{sec/1_intro}

\input{sec/2_related}  
\input{sec/3_method} 
\input{sec/4_experiment}

\input{sec/6_conclusion}


%
%
\bibliographystyle{splncs04}
\bibliography{main}

\end{document}

%% file: sec/0_abstract.tex
\begin{abstract}
We introduce \textbf{\methodname}, a novel framework for semantic-aware 3D shape and texture morphing directly from multi-view images. While 3D Gaussian Splatting (3DGS) enables photorealistic rendering, its unstructured nature often leads to catastrophic geometric fragmentation during morphing. Conversely, traditional mesh-based morphing enforces structural integrity but mandates pristine input topology and struggles with complex appearances. Our method resolves this dichotomy by employing a mesh-guided strategy where a coarse, extracted base mesh acts as a \textit{flexible geometric anchor}. This anchor provides the necessary topological scaffolding to guide unstructured Gaussians, successfully compensating for mesh extraction artifacts and topological limitations. Furthermore, we propose a novel \textit{dual-domain optimization strategy} that leverages this hybrid representation to establish \textit{unsupervised semantic correspondence}, synergizing geodesic regularizations for shape preservation with texture-aware constraints for coherent color evolution. This integrated approach ensures stable, physically plausible transformations without requiring labeled data, specialized 3D assets, or category-specific templates. On the proposed TexMorph benchmark, \methodname substantially outperforms prior 2D and 3D methods, yielding fully textured, topologically robust 3D morphing while reducing color consistency error ($\Delta E$) by 22.2\% and EI by 26.2\%.
 
  \keywords{3D Morphing \and Mesh-guided Deformation  \and Semantic Correspondence \and 3D Gaussian Splatting}
\end{abstract}

%% file: sec/1_intro.tex
\section{Introduction}
\label{sec:intro}
Morphing~\cite{gregory1998feature,zhang2024diffmorpher} has long been a foundational technique in shape transformation, enabling the generation of continuous interpolation sequences between source and target shapes. Serving as a bridge between computer vision and computer graphics, morphing has emerged as an indispensable tool for applications spanning computer animation, geometric modeling, and shape analysis. Its prominence in visual effects for film and media production further underscores its practical significance.

\begin{figure}[t]
  \centering
  \includegraphics[width=\linewidth]{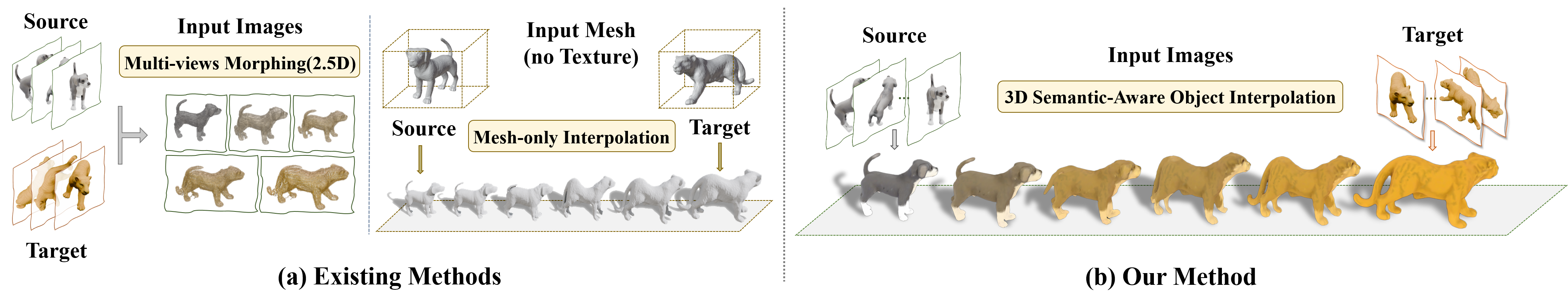}
  \caption{\textbf{ \methodname (b)} achieves fully textured 3D morphing purely from images, overcoming the limitations of prior approaches (a) that either rely on explicit 3D data or lack geometric control. By reconstructing mesh-anchored 3D Gaussians, we employ a \textbf{mesh-guided strategy} to generate semantic-aware intermediate shapes at $t\in[0,1]$.} 
  \label{fig:1}
  \vspace{-2mm}
\end{figure}

Existing morphing techniques can be broadly categorized into two paradigms: image-based methods~\cite{aloraibi2023image,zhang2024diffmorpher} and 3D geometric methods~\cite{eisenberger2021neuromorph,yang2025textured,cao2024spectral}. As summarized in \Cref{fig:1}, these approaches exhibit fundamental trade-offs. Image-based pipelines, such as DiffMorpher~\cite{zhang2024diffmorpher} and FreeMorph~\cite{cao2025freemorph}, produce high-fidelity 2D outputs but lack 3D geometric reasoning and multi-view consistency. Extensions like MorphFlow~\cite{tsai2022multiview} leverage Neural Radiance Fields (NeRF) to address view consistency but are limited by the absence of explicit 3D geometric constraints, resulting in incomplete volumetric reconstructions (denoted as 2.5D* in \Cref{fig:1}). In contrast, 3D-centric methods such as Neuromorph~\cite{eisenberger2021neuromorph} enable mesh-based deformation but require high-quality mesh inputs, neglect texture-aware processing, and struggle with topological complexity. These limitations highlight a critical gap: the lack of a unified framework that balances geometric robustness, textural coherence, and input accessibility without reliance on high-fidelity 3D data.

\begin{wraptable}{r}{0.58\linewidth} 
    \centering
    \vspace{-11mm}
    \caption{The comparison table shows that our method uniquely generates fully textured 3D outputs directly from images, offering complete geometric and textural fidelity.}
    \scriptsize
    \begin{tabular}{cccc}
            \toprule
             Method & Input Type & Output Type & Texture \\  
            \midrule
            DiffMorpher & Images & 2D &\ding{51}  \\ 
            FreeMorph & Images & 2D &\ding{51}  \\
            MorphFlow & Images & 2.5D* &\ding{51} \\  
            Neuromorph  & Mesh & 3D & \ding{55} \\ 
            \methodname (Ours) & Images & 3D & \ding{51}  \\  
            \bottomrule
        \end{tabular}
    \label{table:conparison}
    \vspace{-4mm}
\end{wraptable}

To bridge this gap, 3D Gaussian Splatting (3DGS)~\cite{kerbl20233d} has emerged as a highly capable representation, offering photorealistic rendering directly from images without requiring pre-existing 3D assets. However, pure 3DGS is inherently unstructured. Directly interpolating these discrete Gaussian primitives often leads to catastrophic geometric fragmentation and structural collapse during morphing due to the lack of explicit topological priors. While recent advances~\cite{liu2026interp3d, yin2025wukong, sun2026morphany3d} attempt to incorporate structured latents, they frequently struggle to preserve explicit geometric correspondences. Conversely, simply pipelining traditional mesh morphing onto meshes extracted from 3DGS yields suboptimal results, as it assumes clean topologies and strictly propagates any mesh extraction errors to the final rendering.

In this paper, we propose \textbf{\methodname}, a novel framework that fundamentally rethinks the role of structured priors in point-based radiance fields. Rather than relying on the mesh as a rigid topology constraint, which mandates high-quality inputs and limits flexibility, we introduce a \textbf{mesh-guided strategy} where the underlying extracted topology acts as a \textit{flexible geometric anchor}. This synergy overcomes the limitations of each representation: the coarse mesh provides the necessary scaffold to prevent structural collapse, while the adaptable 3DGS primitives compensate for mesh imperfections and topology limitations to deliver texture-rich rendered outputs.

Central to \methodname is a \textbf{dual-domain optimization strategy} that tightly couples these two representations. By establishing explicit semantic-aware bindings between the flexible mesh anchors and Gaussian primitives, we formulate a deformation process governed by both geodesic-based shape preservation and texture-aware color smoothness. This combined optimization enables unsupervised semantic correspondence without predefined categories, ensuring stable trajectories and seamless temporal coherence entirely from accessible 2D inputs.

Extensive experiments demonstrate that \methodname achieves significant improvements over state-of-the-art methods, robustly handling diverse challenging scenarios including complex topologies and texture-rich objects without relying on high-fidelity 3D inputs. Our main contributions are:
\begin{itemize}
    \item We introduce \methodname, a unified mesh-guided 3DGS framework for semantic-aware morphing that successfully generates high-fidelity 3D transformations directly from images.
    \item We propose a \textit{flexible geometric anchor} formulation that mitigates strict mesh dependency. By relaxing the mesh from a rigid constraint into a dynamic guide, our approach achieves robustness to mesh extraction artifacts and complex topological changes.
    \item We design a novel dual-domain optimization strategy that synergizes unsupervised semantic correspondence, geodesic regularizations, and texture-aware losses, effectively preventing structural fragmentation and surpassing existing methods in geometric and textural coherence.
\end{itemize}

%% file: sec/2_related.tex
\section{Related Work}
\label{sec:Relatedwork}

\subsection{Image Morphing}
Image morphing is a long-standing problem in computer vision and graphics, aiming to generate smooth and perceptually natural transitions between images \cite{aloraibi2023image, zope2017survey, wolberg1998image}. Traditional methods \cite{beier2023feature, bhatt2011comparative, liao2014automating} rely on correspondence-driven warping and blending, which preserve visual consistency but struggle with content creation, often leading to artifacts. More recently, optimal transport has been applied to morphing simple 2D geometries \cite{benamou2015iterative, bonneel2011displacement, solomon2015convolutional}, providing mathematically elegant transformations but lacking the texture richness of natural images. Diffusion-based approaches such as DiffMorpher \cite{zhang2024diffmorpher}, AID \cite{he2024aid}, and FreeMorph \cite{cao2025freemorph} leverage pre-trained generative models to enable flexible morphing across diverse categories.
We instead leverage multi-view inputs to bypass large-scale pre-training and generate mesh-based intermediate representations, enabling shape-aware and texture-consistent 3D morphing.

\subsection{Shape Matching}
\vspace{-2mm}
Establishing point-wise mappings for 3D shape correspondence traditionally relies on geometric constraints \cite{holzschuh2020simulated, roetzer2022scalable} or non-rigid registration \cite{bernard2020mina, eisenberger2019divergence, ezuz2019elastic}, which suffer from costly optimization and limited scalability. Recent learning-based advances address this by matching vertices to templates \cite{monti2017geometric, boscaini2016learning, masci2015geodesic}, leveraging functional maps and diffusion models \cite{litany2017deep, ovsjanikov2012functional, zhuravlev2025denoising}, integrating spatio-spectral cues \cite{cao2024spectral, attaiki2023shape}, or applying 2D priors \cite{liu2025stable}. In contrast, our method eliminates the need for costly 3D inputs and annotations by reconstructing objects and deriving explicit point-wise correspondences purely from images.

\vspace{-2mm}
\subsection{Shape Interpolation}
Shape interpolation addresses the challenge of smoothly transforming one shape into another. Traditional geometric approaches \cite{brandt2016geometric,heeren2012time,wirth2011continuum} formulate this as finding geodesic paths on high-dimensional manifolds, using ARAP \cite{sorkine2007rigid} or PriMo \cite{botsch2006primo} to minimize local distortions. Other strategies include data-driven navigation through shape collections \cite{aydinlilar2021part,gao2017data} and physics-based gradient flows \cite{eisenberger2020hamiltonian,eisenberger2019divergence}, such as MorphFlow's multiview Wasserstein morphing \cite{tsai2022multiview}. Recent neural advances enable robust unsupervised interpolation \cite{eisenberger2021neuromorph}, with spatio-spectral integrations effectively handling large non-isometric deformations \cite{cao2024spectral}. Furthermore, while methods leveraging 2D correspondence \cite{liu2025stable} or diffusion priors \cite{yang2025textured} show promise, they often struggle to maintain explicit geometric consistency.

\subsection{Textured 3D Morphing}
Recently, generative models have emerged to jointly synthesize 3D geometry and appearance, utilizing diffusion priors~\cite{yang2025textured}, flow-based optimal transport~\cite{yin2025wukong}, and Structured Latent  blending~\cite{sun2026morphany3d, liu2026interp3d}. While these paradigms powerfully hallucinate visually plausible intermediate states via latent space interpolation, their unconstrained synthesis inherently lacks explicit 3D topological guarantees. In contrast, rather than relying on unpredictable latent priors, we explicitly anchor 3D Gaussians to a mesh manifold. By enforcing geodesic and ARAP constraints, our framework strictly preserves physical trajectories and local rigidity, ensuring geometrically sound and texture-consistent morphing.

\begin{figure*}[t]
  \centering
  \includegraphics[width=1\textwidth]{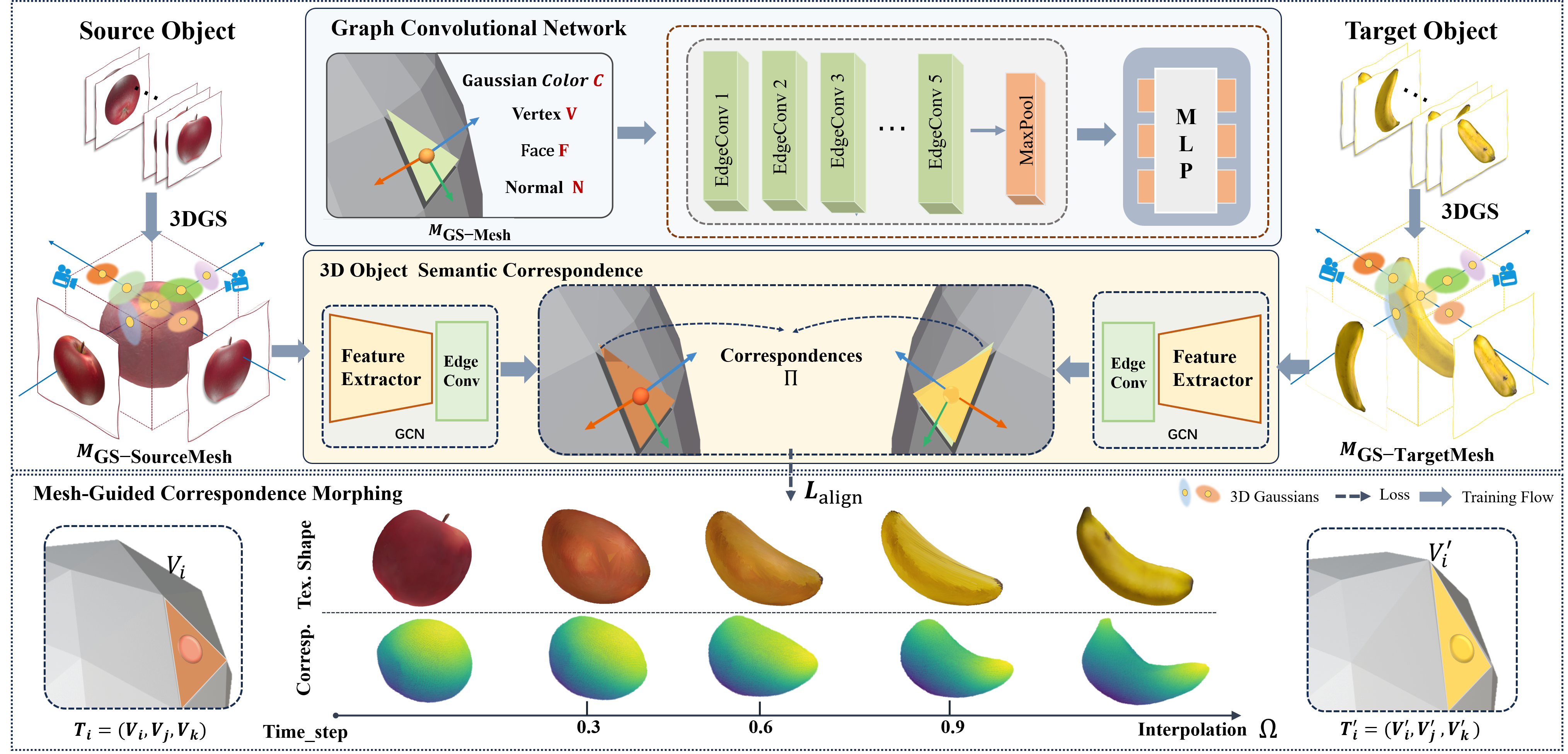}
\caption{\textbf{\methodname} performs high-quality 3D morphing between source $\mathcal{X}$ and target $\mathcal{Y}$ images. We extract surface meshes from 3D Gaussian Splatting (Sec.~\ref{sub:gaussion}) to align geometry and texture. Based on the correspondence matrix $\Pi_{XY}$ (Sec.~\ref{sub:Correspondence}), intermediate shapes are interpolated and optimized via a joint loss (Sec.~\ref{sub:loss}). (Top: Blender renderings; bottom: Matplotlib-based correspondence visualizations.)}
  \label{fig:pipeline}
  \vspace{-1mm}
\end{figure*}

%% file: sec/3_method.tex
\section{Mesh-Guided Gaussian Morphing}
\vspace{-2mm}

Given source and target objects represented by multi-view images, we propose a semantic-aware 3D morphing framework that addresses a fundamental challenge: achieving geometrically consistent transformations while preserving photorealistic surface details. The core problem is that modern explicit representations present a trade-off: 3D Gaussian Splatting (3DGS) \cite{kerbl20233d} lacks the topological connectivity needed for structured morphing, while traditional meshes struggle to model complex appearance.

As shown in \Cref{fig:pipeline}, Our framework, Mesh-Guided Gaussian Morphing, resolves this tension by introducing a novel hybrid paradigm. Our key insight is to impose an explicit triangular mesh as a \emph{topological scaffold} to guide the transformation of unstructured Gaussians. By anchoring Gaussians to this mesh, we can leverage powerful mesh-based correspondence techniques to establish semantic connections. This allows us to compute a geometrically consistent morphing flow in the structured mesh domain while using the rich Gaussian representation for photorealistic rendering at any point in the transformation.

\subsection{Hybrid Mesh-Gaussian Representation for Semantic Morphing}
\label{sub:gaussion}
\noindent \textbf{The Connectivity Challenge in Gaussian-Based Morphing.} 3DGS represents a scene as a set of anisotropic 3D Gaussians, which can be optimized to reproduce a set of input images, enabling photorealistic novel-view synthesis. Each Gaussian $g$ is defined by its position $\mu_g \in \mathbb{R}^3$, covariance $\Sigma_g$, opacity $\alpha_g$, and spherical harmonics (SH) coefficients $\mathbf{sh}_g$. While excellent for rendering, the discrete, unstructured nature of these Gaussians prevents the establishment of meaningful semantic correspondences between objects. A direct Gaussian-to-Gaussian matching would likely produce geometrically implausible results that tear or distort the structure of the object.

\noindent \textbf{Mesh-Anchored Gaussian Binding.} To overcome the unstructured nature of pure 3DGS and prevent catastrophic fragmentation during morphing, we introduce a mesh-anchored representation. This explicit topological scaffold provides the discrete Gaussian primitives with a rigorous geometric anchor.

First, rather than demanding a perfect geometric prior, we distill a base triangular mesh from the initial optimized 3DGS radiance field. Building upon established surface-alignment principles~\cite{guedon2024sugar, guedon2024gaussian}, we employ regularized Poisson reconstruction~\cite{kazhdan2006poisson} to extract a topological blueprint. Crucially, as highlighted in \Cref{sec:intro}, this extracted mesh does not need to be flawlessly high-fidelity. Instead, it serves as a \textit{flexible geometric anchor} a coarse but structurally sound scaffold that guides the subsequent deformation while the adaptable Gaussians compensate for any local extraction artifacts.

This mesh scaffold establishes a differentiable binding between discrete Gaussians and the continuous surface manifold. Each Gaussian is uniquely anchored to a triangular face $f=(V_1, V_2, V_3)$ with normal $\mathbf{n}_f$, its position $\mu_g$ parameterized by barycentric coordinates $(w_1, w_2, w_3)$ and a normal offset $d$:
\begin{equation}
\mu_g = w_1 V_1 + w_2 V_2 + w_3 V_3 + d \cdot \mathbf{n}_f.
\label{eq:binding}
\end{equation}
This binding ensures that as the mesh vertices $V_i$ deform over the course of the morph, the anchored Gaussians move cohesively with the surface, preserving the fine-grained geometric and appearance details they represent.

\subsection{Unsupervised Topology-Aware Semantic Correspondence}
\label{sub:Correspondence}
\noindent \textbf{Semantic-Aware Mesh Connectivity.} With the flexible geometric anchor established, we tackle the core challenge of identifying corresponding regions between the source and target objects. Unlike traditional methods requiring predefined keypoints, fully isometric topologies, or category-specific templates, we formulate this as an \textit{unsupervised} correspondence problem between the source mesh $(V^S, F^S)$ and target mesh $(V^T, F^T)$. By leveraging a hybrid graph representation (detailed in \Cref{sub:loss}), this module gracefully bridges moderate topological gaps between objects. The correspondence is encoded as a probabilistic matrix $\Pi \in \mathbb{R}^{n \times m}$:
\begin{equation}
\Pi_{ij} = P(V^T_j \mid V^S_i) = \frac{\exp(\sigma c_{ij})}{\sum_{k=1}^{m} \exp(\sigma c_{ik})},
\label{eq:prob_dist}
\end{equation}
where $c_{ij}$ is the cosine similarity between learned features of source $V^S_i$ and target $V^T_j$. To learn semantically rich features, we use a 5-layer Graph Convolutional Network (GCN) that processes mesh connectivity, allowing it to capture local geometric context without relying on hand-engineered descriptors.

\noindent \textbf{Neural Morphing Flow.} Rather than simple linear interpolation, we learn a continuous, non-linear deformation field. We employ a neural network, the Correspondence Morphing Flow ($\Psi$), to predict the morphing trajectory. This network ($\Psi$) utilizes the same GCN architecture as our correspondence network, but additionally accepts the time $t$ as an input. This allows $\Psi$ to learn a continuous, non-linear interpolation flow by conditioning its predicted displacement on the time $t \in [0, 1]$. At any time $t \in [0,1]$, the morphed source vertices $V^S(t)$ are given by:
\begin{equation}
V^S(t) = V^S + \Psi(V^S, \Pi V^T - V^S, t).
\label{eq:morphing}
\end{equation}
Here, the term $\Pi V^T - V^S$ represents the \emph{semantically-aligned displacement field} that maps each source vertex to its corresponding target location. The network $\Psi$ learns to smoothly interpolate this displacement over time. 

\noindent \textbf{Consistent Gaussian Updates.} As the mesh vertices $V^S(t)$ deform, the positions of the bound Gaussians $\mu_g(t)$ are updated consistently via the barycentric relationship established in Eq. \ref{eq:binding}:
\begin{equation}
\mu_g(t) = \sum_{i=1}^{3} w_i V_{f_i}(t),
\end{equation}
where $V_{f_i}(t)$ are the deformed positions of the vertices of the triangle $f$ to which Gaussian $g$ is bound. This maintains the tight coupling between the mesh and the Gaussians throughout the entire morphing sequence. To evolve appearances, time-dependent SH coefficients are dynamically optimized via smoothness loss gradients (\Cref{sub:loss}) towards linearly interpolated target colors. Opacities and scales are linearly interpolated over time $t$.


\subsection{Dual-Domain Optimization for Plausible Morphing}
\label{sub:loss}
To prevent structural fragmentation and ensure seamless visual transitions, we optimize the correspondence matrix $\Pi$ and the morphing flow network $\Psi$ using our proposed \textbf{dual-domain optimization strategy}. This unified objective balances constraints in both the geometric domain (geodesic structure and spatial alignment) and the texture domain (appearance consistency).

\noindent \textbf{Geometric Consistency.} To prevent unnatural stretching and distortion, we enforce that the intrinsic geometric structure of the surfaces is preserved. We measure this using geodesic distances on the mesh. To compute the geodesic distance $D_{\text{g}}(i, j)$ between any two vertices, we run Dijkstra's algorithm on a hybrid graph formed by the union of the mesh adjacency graph $G_{\text{adj}}$ (preserving topology) and a KNN graph $G_{\text{knn}}$ (adding shortcuts to better approximate Euclidean distances). Further details are provided in Appendix A.2. The geodesic distortion loss is then:
\begin{equation}
\mathcal{L}_{\text{geo}} = \big\| \Pi D_{\text{g}}^T \Pi^\top - D_{\text{g}}^S \big\|_F^2,
\label{eq:geoloss}
\end{equation}
where $D_{\text{g}}^S$ and $D_{\text{g}}^T$ are the geodesic distance matrices for the source and target meshes, and $\|\cdot\|_F$ is the Frobenius norm. This loss encourages the correspondence $\Pi$ to map regions of the target mesh back to the source mesh in a way that respects their intrinsic geometry.

To further encourage local rigidity, we add an As-Rigid-As-Possible (ARAP) energy term~\cite{sorkine2007rigid}, which penalizes non-rigid deformations. We evaluate this over sampled timesteps during the morph:
\begin{equation}
\mathcal{L}_{\text{arap}} = \mathbb{E}_{t \sim U[0,1]} \left[ E_{\text{arap}}(\mathbf{X}(t), \mathbf{X}(t+\delta t)) \right],
\label{eq:arap}
\end{equation}
where $\mathbf{X}(t)$ is the mesh state at time $t$ and $\delta t$ is a small perturbation.

\noindent \textbf{Texture-Aware Appearance Consistency.} To ensure smooth visual transitions in the texture domain, we introduce a texture-aware geodesic smoothness loss on the vertex colors. We first initialize the color \(C^i\) of each vertex \(i\) by averaging the RGB colors of its bound Gaussians (with SH coefficients evaluated from a canonical viewing direction). To encourage color coherence during deformation, the loss penalizes color differences between adjacent vertices, weighted inversely by their geodesic distance:
\begin{equation}
\mathcal{L}_{\text{smooth}} = \sum_{(i,j) \in E_{\text{adj}}} \frac{1}{D_{\text{g}}(i,j) + \epsilon} \cdot \big\| C^i(t) - C^j(t) \big\|_2^2,
\label{eq:smoothloss}
\end{equation}
where $C^i(t)$ represents the color descriptor of vertex $i$ at time $t$, and $E_{\text{adj}}$ is the set of edges in the mesh adjacency graph. The smoothing loss incorporates a small constant $1 \times 10^{-5}$ to ensure numerical stability. This encourages coherent texture evolution while allowing for sharp semantic boundaries across distant parts of the object.
Crucially, gradients from $\mathcal{L}_{smooth}$ directly backpropagate to update the time-dependent SH coefficients of the anchored Gaussians, enforcing smooth color interpolation and obviating the need for intermediate multi-view photometric supervision.

\noindent \textbf{Semantic Alignment Constraint.} To ensure the morphing sequence reaches its destination, we add a terminal constraint that drives the deformed source mesh to the target configuration at the final timestep:
\begin{equation}
\mathcal{L}_{\text{align}} = \big\| V^S(t=1) - \Pi V^T \big\|_F^2.
\label{eq:align}
\end{equation}
This loss acts as a boundary condition, ensuring that the morph respects the learned semantic correspondences.

\noindent \textbf{Unified Loss Function.} Our final objective function is a weighted sum of these components:
\begin{equation}
\mathcal{L}_{\text{total}} = \lambda_{\text{geo}} \mathcal{L}_{\text{geo}} + \lambda_{\text{arap}} \mathcal{L}_{\text{arap}} + \lambda_{\text{smooth}} \mathcal{L}_{\text{smooth}} + \lambda_{\text{align}} \mathcal{L}_{\text{align}},
\label{eq:total_loss}
\end{equation}
where the $\lambda$ hyperparameters balance the competing objectives of geometric fidelity, structural rigidity, appearance consistency, and semantic alignment.

%% file: sec/4_experiment.tex
\section{Experiments}
\begin{figure*}[t]
  \centering
  \includegraphics[width=1\textwidth]{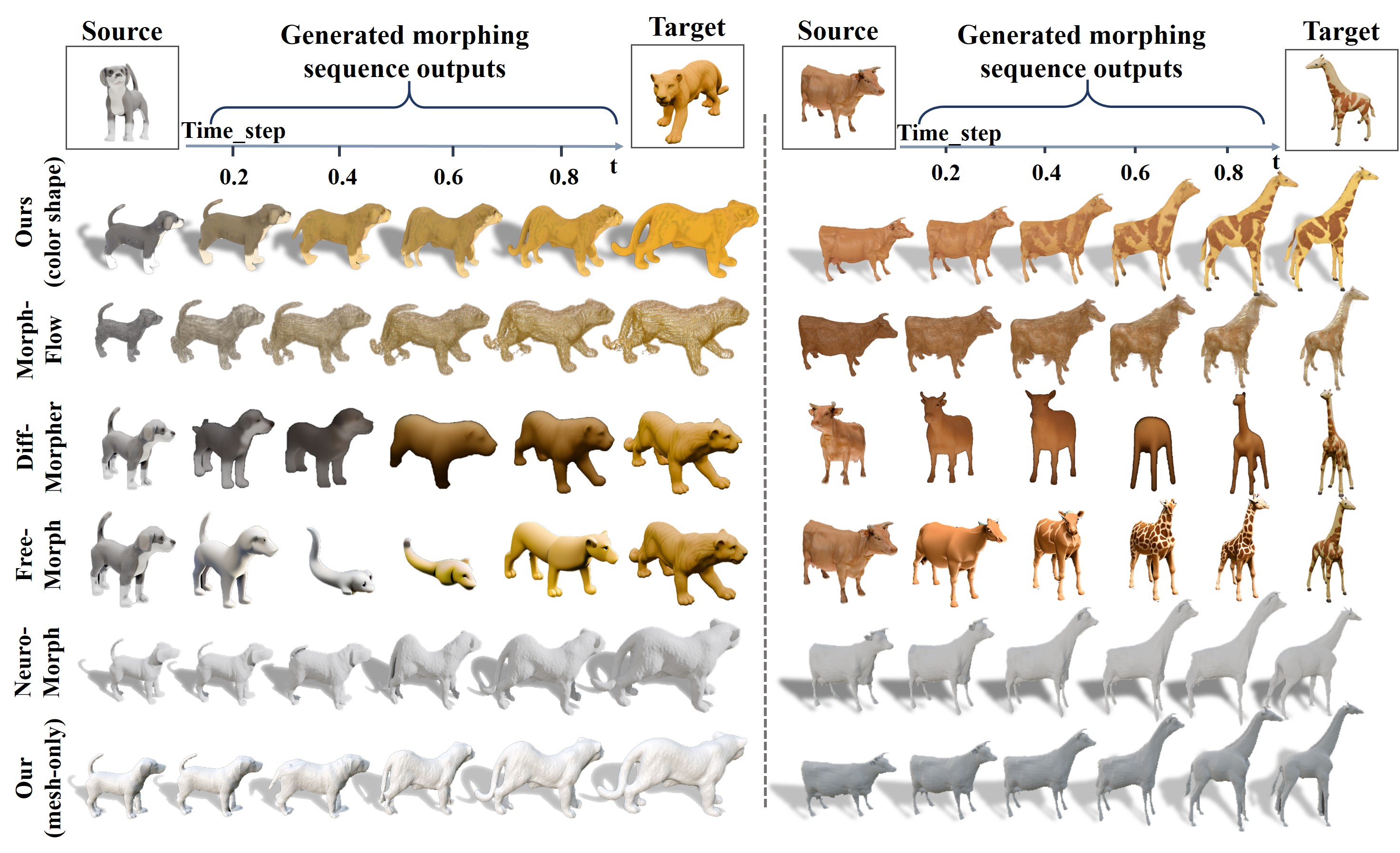}
  \caption{Qualitative comparison of morphing methods on the benchmark dataset. 
Baselines include DiffMorpher~\cite{zhang2024diffmorpher} and FreeMorph~\cite{cao2025freemorph} for image morphing, 
NeuroMorph~\cite{eisenberger2021neuromorph} for texture-free 3D shape morphing, 
and MorphFlow~\cite{tsai2022multiview} for textured multi-view morphing without true geometry. 
Our method generates textured 3D morphing with geometric details directly from image inputs.}
  \label{fig:Qualitatively result}
  \vspace{-3mm}
\end{figure*}

We conduct comprehensive experiments to demonstrate the efficacy of our method in producing high-quality, semantically consistent 3D morphs. To this end, we introduce TexMorph, a novel benchmark tailored specifically for this task. Our evaluation protocol encompasses rigorous quantitative comparisons against state-of-the-art 2D and 3D baselines using novel metrics, qualitative assessments of the generated morphing trajectories, and a detailed ablation study to validate the contributions of our core components.

\subsection{Experimental Setup} \label{sec:setup}

\noindent\textbf{TexMorph Benchmark.}
To rigorously evaluate 3D morphing from multi-view images, we created a new benchmark named \textbf{TexMorph} (\textbf{T}exture-rich, \textbf{M}orphing-focused). The benchmark is comprised of challenging source-target pairs designed to test geometric and appearance transformations. It includes: (1) high-fidelity synthetic models with complex textures rendered from multiple viewpoints; (2) real-world objects captured via 3D scanning; and (3) objects captured in-the-wild using standard mobile phone cameras. The dataset features over ten object categories, including animals, fruits, and furnitures, providing diverse topological and textural challenges. See Appendix A.1 for details.

\begin{figure}[t]
  \centering
  \includegraphics[width=\linewidth]{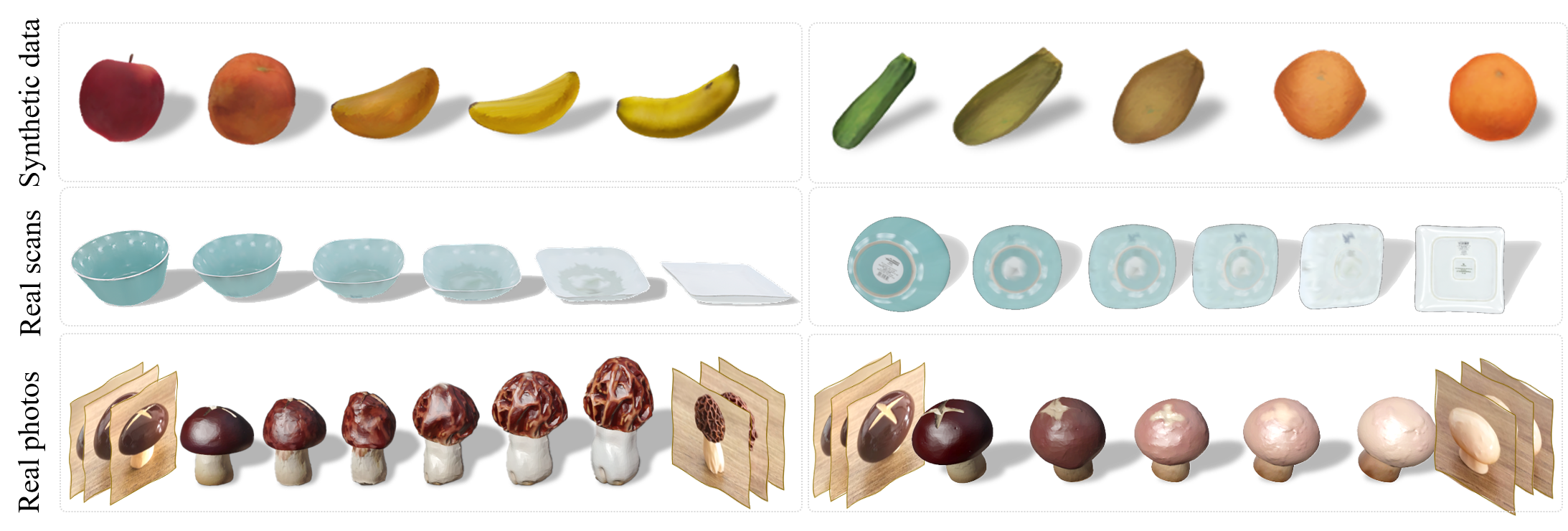}
   \caption{Qualitative morphing results with non-isometric deformations, demonstrating robust interpolation under challenging geometric conditions (Up: synthetic datas; Middle: real-world scanned objects from GSO \cite{downs2022google}; Bottom: real-world photos).}
   \vspace{-2mm}
   \label{fig:non-isometric}
\end{figure}

\noindent\textbf{Evaluation Metrics.} 
Standard novel-view synthesis metrics are inadequate for evaluating dynamic morphing. We thus propose three specialized metrics to assess the spatio-temporal quality from source ($t=0$) to target ($t=1$):

\begin{description}
    \item[$\bullet$ Structural Stability (MSE-SSIM):] 
    Measures geometric consistency by computing the Mean Squared Error of the actual SSIM against an ideal linear trajectory: $\mathcal{E} = \frac{1}{N} \sum_{t \in T} \sum_{I \in \{A, B\}} ( \text{SSIM}_{\text{ideal}} - \text{SSIM}_{\text{actual}} )^2$. Lower values indicate higher stability with fewer structural artifacts.

    \item[$\bullet$ Color Consistency ($\Delta E$):] 
    Assesses appearance smoothness by calculating the CIELAB perceptual color difference between corresponding surface points, where $\Delta E_{ab}^* = \sqrt{(\Delta L^*)^2 + (\Delta a^*)^2 + (\Delta b^*)^2}$. Lower $\Delta E$ signifies a smoother transition without color bleeding.

    \item[$\bullet$ Edge Integrity (EI):] 
    Evaluates silhouette continuity by measuring the stability of rendered edge maps: $EI = \text{Count}(\text{Canny}(I, \tau_{l}, \tau_{h})) - 1$. Lower scores indicate less fragmented edges and more stable structural transitions.
\end{description}
\vspace{-1mm}

Detailed formulations are available in Appendix A.3.

All experiments were conducted on 2 NVIDIA RTX A6000 GPU. Generating the initial hybrid mesh-Gaussian representation takes approximately 1 hour per object pair as a pre-processing step, followed by model training that requires 500 to 1000 iterations. While this initial setup is more time-consuming than tuning-free 2D methods, it acts as a fully amortized investment: the optimized model exhibits strong intra-category generalization. This effectively bypasses the need for repetitive per-pair retraining. Consequently, inferring and rendering a full, high-resolution 3D morphing sequence for novel object pairs within the same category takes merely 2 minutes, ensuring  efficient downstream generation.

\subsection{Evaluation}\label{sec:Comparison}

\noindent We perform a comprehensive evaluation of SemMorph3D against several SOTA 2D and 3D morphing methods. For 2D baselines, we compare against DiffMorpher \cite{zhang2024diffmorpher}, a diffusion-based method, and FreeMorph \cite{cao2025freemorph}, a tuning-free approach. For 3D baselines, we include MorphFlow \cite{tsai2022multiview}, which leverages optimal transport for multi-view transitions, NeuroMorph \cite{eisenberger2021neuromorph}, which computes topology-aligned shape correspondences, and Interp3D \cite{liu2026interp3d}, a training-free generative framework that interpolates in the structured latent space.

\noindent\textbf{Textured Morphing Analysis.}
Our method excels at producing smooth, high-fidelity texture transitions, as qualitatively demonstrated in Figure~\ref{fig:Qualitatively result}. The linear color interpolation of MorphFlow is inadequate for high-dimensional color spaces, leading to oversmoothed transitions and loss of detail. For example, during the ``dog$\to$lion" transformation, it reduces the morph to a simple color shift, failing to preserve the intricate fur patterns of the lion or the distinct white patches of the dog. The 2D methods perform poorly on challenging cross-category pairs; DiffMorpher fails in both geometric and color alignment, while the recent tuning-free 2D SOTA, FreeMorph, struggles to maintain consistent 3D structures across views, introducing severe structural artifacts (unnatural lizard-like textures) and color oversaturation. While tuning-free 2D methods offer faster initial generation, they entirely lack explicitly controllable 3D geometry. In contrast, our approach explicitly coordinates 3D spatial properties and achieves superior color fidelity, corroborated by lower $\Delta E$ values (Table~\ref{table:metric}), and maintains fine-grained texture details throughout the transformation.
Furthermore, \Cref{fig:non-isometric} illustrates that our method yields visually coherent and plausible morphing results even under non-isometric deformations, demonstrating its robustness against challenging geometric conditions. By covering synthetic models, real-world scanned objects from GSO \cite{downs2022google}, and photographs of everyday items, the results further highlight the generalization ability of Gaussian morphing across diverse data sources.



\begin{figure}[t]
    \centering
    \begin{minipage}{0.455\linewidth}
        \includegraphics[width=\linewidth]{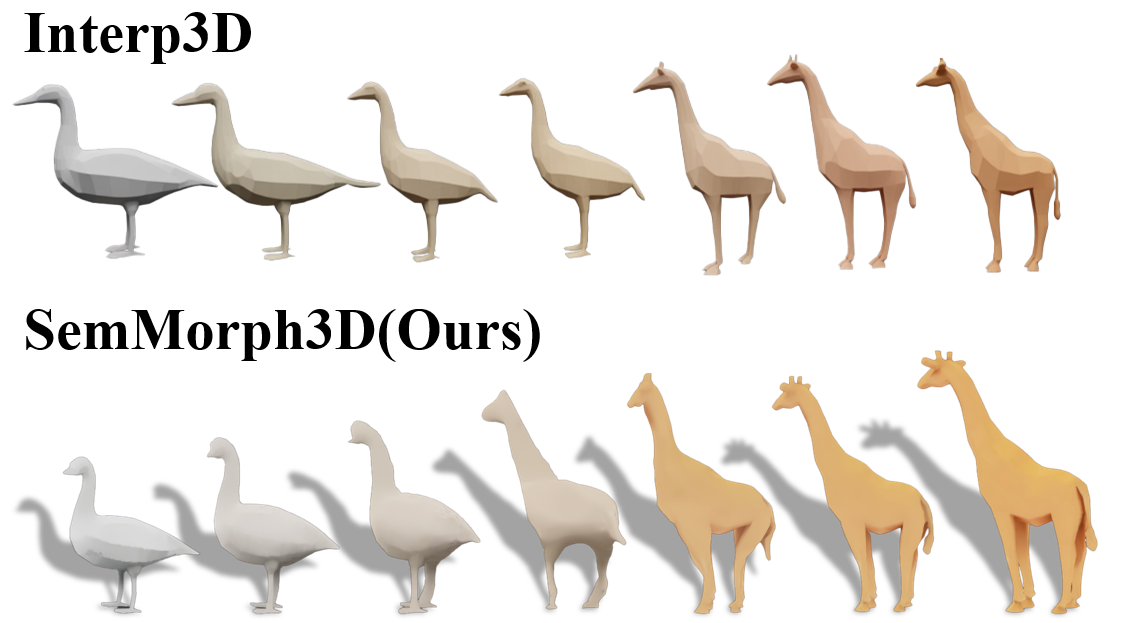}
        \caption{Qualitative comparison of 3D morphing results on the ``goose-to-giraffe'' pair between Interp3D \cite{liu2026interp3d} and our proposed method.}
        \label{fig:compare}
    \end{minipage}
    \hfill
    \begin{minipage}{0.50\linewidth}
        \centering
        \vspace{2mm}
        \includegraphics[width=0.98\linewidth]{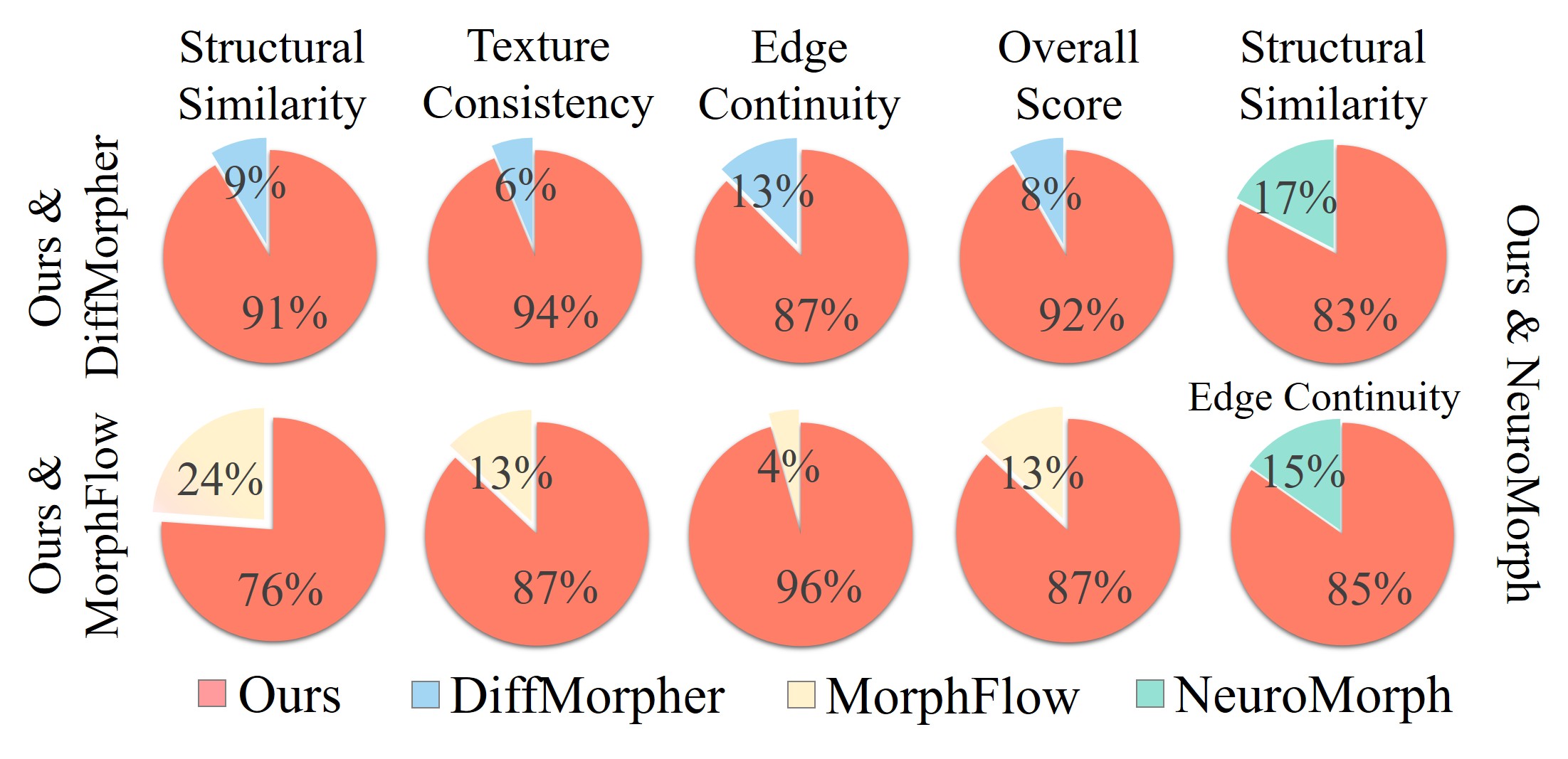}
        \caption{User study results comparing our method with baselines. Our approach is consistently preferred for color consistency, structural similarity, and edge continuity.}
        \label{fig:userstudy1}
    \end{minipage}
    \vspace{-3mm}
\end{figure}

\noindent\textbf{Geometric and Structural Analysis.}

\begin{wraptable}{r}{0.45\linewidth} 
    \centering
    \footnotesize
    \vspace{-11mm}
    \caption{Quantitative comparison of structural similarity (MSE of SSIM), color consistency ($\Delta E$), and edge continuity (EI).}
    \begin{tabular}{cccc}
        \toprule
        Method & MSE(SSIM)$\downarrow$ & $\Delta E$$\downarrow$ & EI $\downarrow$ \\
        \midrule
        DiffMorpher  &0.19& 105 & 97  \\   
        MorphFlow  & 0.17&8.23 & 33.6  \\  
        Neuromorph  & 0.13&/ & 13.0  \\ 
        FreeMorph  & 0.20 &13.0 & 21.6 \\ 
        Our & \textbf{0.11}&\textbf{6.40} & \textbf{9.0}  \\ 
        \bottomrule
    \end{tabular}
    \vspace{-6mm}
    \label{table:metric}
\end{wraptable}

As shown in Table~\ref{table:metric}, our method achieves state-of-the-art structural consistency and edge continuity, primarily due to $\mathcal{L}_{geo}$, which preserves local geometric details. For a fair comparison with NeuroMorph, we use the same input meshes for both methods. NeuroMorph relies on mesh connectivity for geodesic computation, making it brittle when handling fragmented or coarse geometries. Our hybrid graph representation bypasses this dependency, yielding a more robust and efficient solution. Furthermore, our semantic-aware mechanism produces more plausible deformations, correctly preserving features like the tail in ``dog$\to$lion" morphs and neck details in giraffe morphs, where NeuroMorph falters. MorphFlow suffers from a lack of constraints on mesh topology or semantic information, an absence that leads to noticeable edge fragmentation and silhouette tearing. Although Interp3D \cite{liu2026interp3d} achieve visually plausible transitions across severe topological gaps (e.g., the ``goose$\to$giraffe" morph in \Cref{fig:compare}), they inherently rely on latent interpolation without explicit 3D point-wise correspondences. By contrast, our topology-aware framework effectively avoids these issues by leveraging the mesh structure. This ensures that the structural evolution is governed by physically interpretable local rigidity and enhanced edge continuity, rather than unconstrained generative hallucination.

\begin{figure}[t]
  \centering
  \begin{minipage}[t]{0.48\textwidth}
    \centering
    \includegraphics[width=\linewidth]{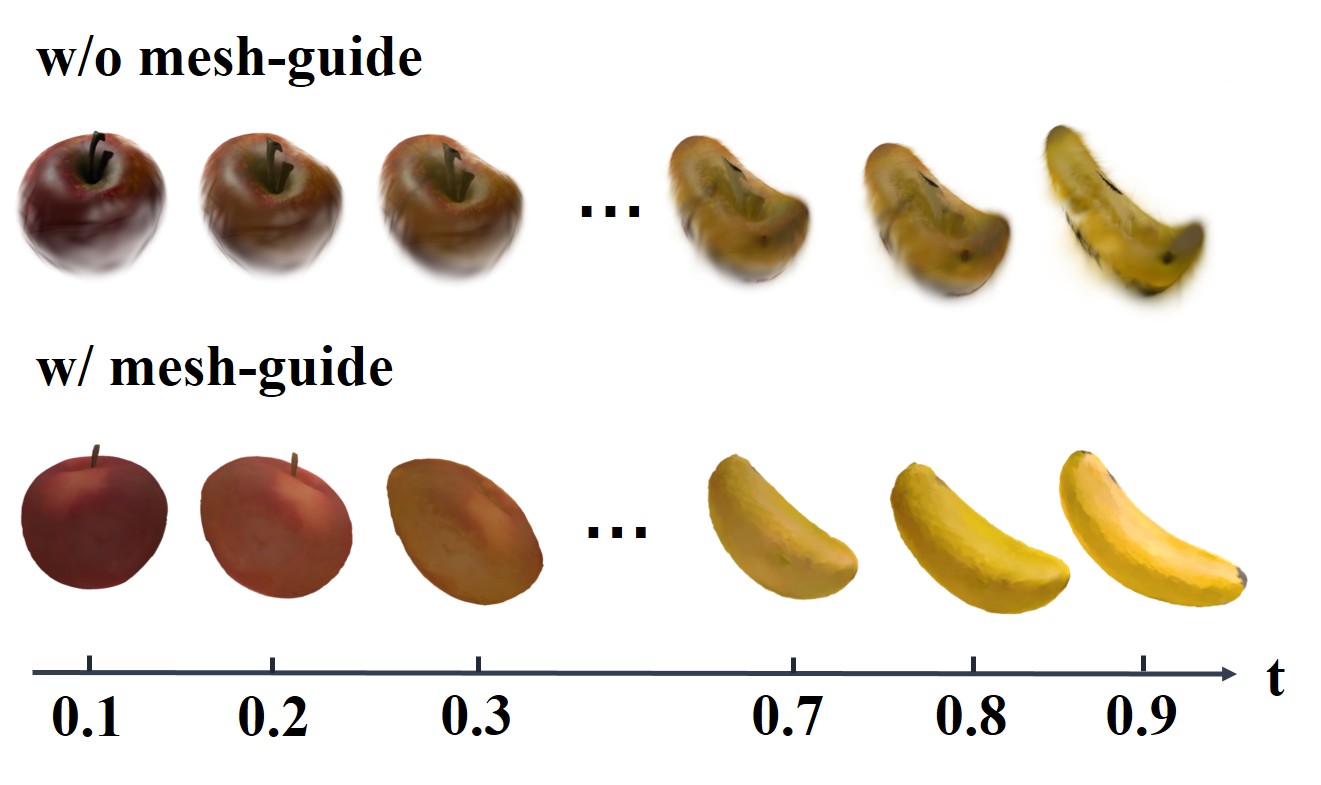}
    \caption{\textbf{Ablation on mesh-guided strategy.} Top: Morphing without our strategy. Bottom: Our approach, enabling edge-continuous and smooth transitions.}
    \label{fig:Ablation}
  \end{minipage}
  \hfill
  \begin{minipage}[t]{0.48\textwidth}
    \centering
    \includegraphics[width=\linewidth]{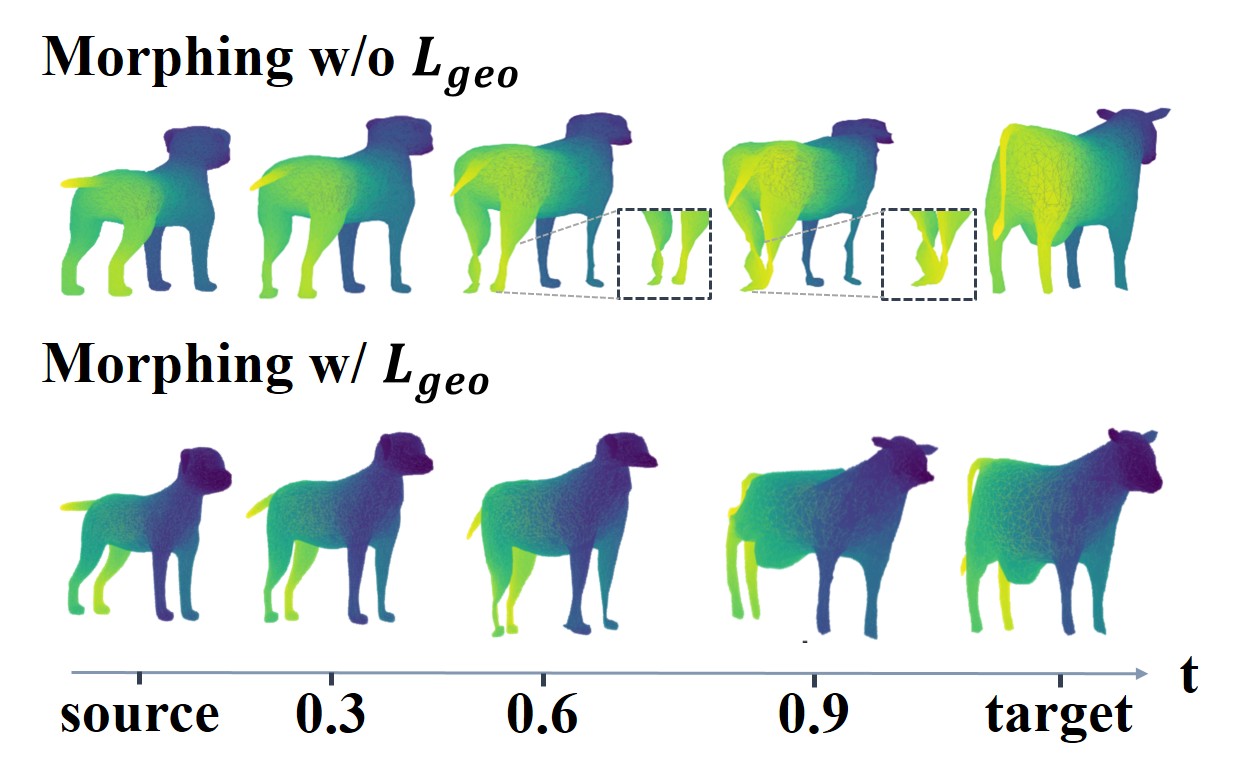}
    \caption{Ablation Study for \textbf{geometric distortion loss}. Comparison of morphing results without (up) and with (below) the geometric distortion loss.}
    \label{fig:Ablationloss}
  \end{minipage}
\vspace{-3mm}
\end{figure}

\noindent\textbf{User Study.}
To validate the perceptual quality of our results, we conducted a user study with 54 participants, who compared outputs of our method against those from DiffMorpher, MorphFlow, NeuroMorph and the Ablation study. The evaluation focused on four criteria: structural similarity, texture consistency, edge continuity, and overall preference.The criteria shown below:

\begin{itemize}
    \item \textbf{Structural Similarity}: Preservation of structure in intermediate frames.
    \item \textbf{Texture Consistency}: Smooth, natural transitions without abrupt jumps.
    \item \textbf{Edge Continuity}: Smooth, continuous edges without breaks or distortions.
    \item \textbf{Overall Score}: Weighted evaluation of structure, texture, and edge metrics.
\end{itemize}

Full details are provided in Appendix A.5. The results show an overwhelming preference for our method across all metrics. Over $80\%$ of users rated our morphs as superior overall, with particularly strong and consistent agreement on aspects such as texture consistency and edge continuity. This perceptual validation confirms that our method generates more visually coherent and high-quality morphs, aligning with our quantitative experiments.



\subsection{Ablation Study}\label{sec:Ablation}

We conducted an ablation study to isolate the contributions of our core components: the mesh-guided strategy and the geometric distortion loss.

\noindent\textbf{Importance of Mesh Guidance.}

To evaluate the critical role of mesh guidance in maintaining morphing coherence, we first conducted a comparative evaluation between two distinct approaches: (1) a variant of our method that removes mesh guidance, relying solely on point-based morphing, and (2) our full model, which incorporates mesh guidance by leveraging the complete mesh topology (including vertices, edges, faces, and normals) to establish a shared correspondence $\Pi$. As summarized in Table~\ref{table:Ablation} and illustrated in Figure~\ref{fig:Ablation}, the point-based variant fails to maintain structural coherence, resulting in significant tearing and discontinuities along object surfaces, particularly noticeable at edges.Quantitatively, this structural degradation is reflected in a higher Edge Continuity Index (EI) score of 34.3, indicating poorer performance. The absence of topological guidance also compromises texture quality, leading to blurry artifacts. In contrast, the mesh-guided approach effectively preserves structural integrity and ensures smoother transitions by enforcing spatial and textural consistency through the explicit use of mesh structure.


\noindent\textbf{Role of Geometric Distortion Loss.} 

\begin{wraptable}{r}{0.6\linewidth} 
    \centering
    \footnotesize
    \vspace{-11mm}
    \caption{Mesh-Guided Strategy Ablation: Quantifying edge continuity (EI), 
    user-rated transition quality, and texture preservation (MSE(SSIM)) to validate 
    the importance of mesh guidance for smooth shape and texture morphing.}
    \begin{tabular}{@{}lcccl@{}}
        \toprule
        & \multicolumn{2}{c}{Edge Continuity} 
        & \multicolumn{1}{c}{Texture Quality} \\
        \cmidrule(lr){2-3}
        \cmidrule(lr){4-5}
        & EI$\downarrow$ & User$\uparrow$ 
        & \multicolumn{2}{c}{MSE(SSIM)$\downarrow$} \\
        \midrule
        w/o Mesh-Guided  & 34.3 & 0.02 &  0.34 & \\
        w/o $\mathcal{L}_{smooth}$ & -- & -- & 0.22 & \\
        Ours & \textbf{9.0} & \textbf{0.98} & \textbf{0.11} & \\
        \bottomrule
    \end{tabular}
    \vspace{-6mm}
    \label{table:Ablation}
\end{wraptable}
Next, we ablate the geometric distortion loss. Without this constraint, the morphing process introduces severe and unnatural deformations, such as the distorted leg geometry shown in Figure~\ref{fig:Ablationloss}. These artifacts not only degrade visual quality but also disrupt the structural plausibility of the interpolated shapes,making the transitions appear unrealistic. By explicitly penalizing local shape changes, this loss serves as a key regularizer that preserves structural integrity, enforces geometric continuity, and produces smoother, more plausible transformations. User feedback corroborates this finding, confirming a marked reduction in visual distortion when the loss is applied.

\noindent\textbf{Role of Texture-Aware Smoothness Loss.} 

To address concerns that the appearance consistency loss ($\mathcal{L}_{smooth}$) might induce over-smoothing, we evaluate its exclusion in Table~\ref{table:Ablation}. Without $\mathcal{L}_{smooth}$, the morph suffers from high-frequency color artifacts and texture bleeding, exacerbating texture degradation (MSE-SSIM: 0.22 vs.\ 0.11). Crucially, because $\mathcal{L}_{smooth}$ is inversely weighted by geodesic distance, it restricts the smoothing effect to local semantic patches. This effectively preserves sharp texture boundaries across distant geometric regions, circumventing global blur.

\noindent\textbf{Impact of ARAP Loss on Structural Integrity.} 

To verify the contribution of the As-Rigid-As-Possible (ARAP) constraint, we visualize the morphing sequence with this term disabled. Fundamentally, the ARAP loss serves as a temporal regularizer, computed between adjacent frames ($t$ and $t+\delta t$) throughout the optimization process. By penalizing non-rigid deformations across these consecutive time steps, it enforces the preservation of local geometric features, ensuring that the mesh triangles undergo physically plausible rotations and translations rather than arbitrary scaling or shearing. 
\begin{figure}[t]
  \centering
  \begin{minipage}[h]{0.43\textwidth}
    \centering
    \includegraphics[width=\linewidth]{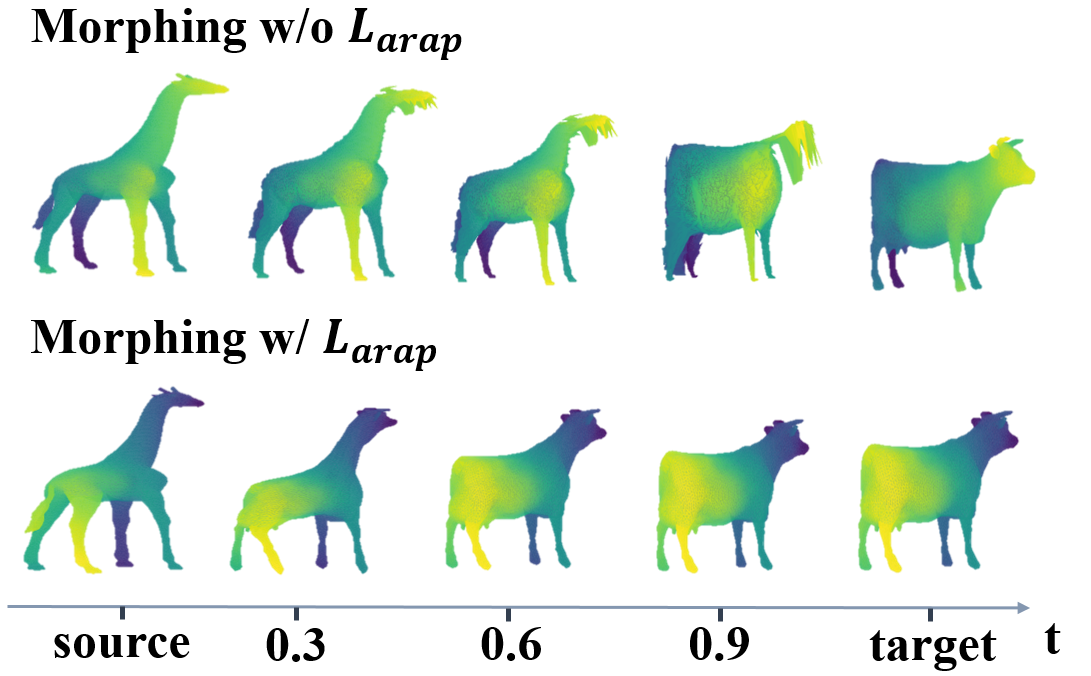}
    \caption{Ablation Study for \textbf{ARAP loss}. Comparison of morphing results without (top) vs. with (bottom).}
    \label{fig:Ablationarap}
  \end{minipage}
    \hfill
 \begin{minipage}[h]{0.53\textwidth}
    \centering
    \includegraphics[width=\linewidth]{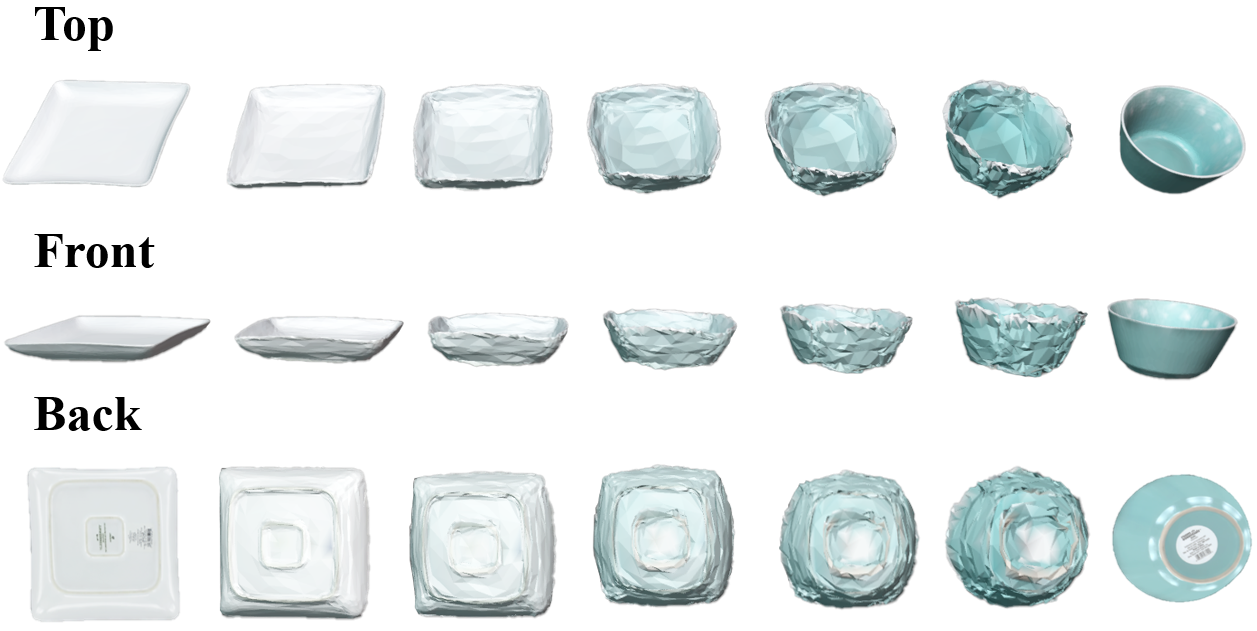}
    \caption{\textbf{ARAP Ablation.} Multi-view visualization of the ``plate-to-bowl'' morph without ARAP loss.}
    \label{fig:Ablationaraploss}
  \end{minipage}
  \vspace{-2mm}
\end{figure}
As observed in our experiments (see in Figure~\ref{fig:Ablationaraploss} and Figure~\ref{fig:Ablationarap}), removing this constraint leads to unnatural distortions; specifically, the ``plate-to-bowl'' case suffers from a loss of structural integrity, while the ``giraffe-to-cow'' sequence exhibits severe edge twisting and arbitrary stretching.

In summary, these studies demonstrate that the synergy between mesh guidance, geometric distortion loss and the ARAP loss is essential for achieving high-fidelity geometric and textural transformations, significantly improving geometric continuity and leading to more natural morphing results.

%% file: sec/6_conclusion.tex
\section{Discussion and Conclusion}
\label{sec:discussion}

\textbf{Discussion and Limitations.} 
While SemMorph3D achieves high-fidelity and texturally coherent 3D morphing, it inherits certain limitations. Primarily, geometric accuracy relies heavily on the base mesh extracted from the initial 3DGS field. For extremely thin structures or complex topologies, rigid mesh-guided anchors may limit local flexibility, causing minor visual artifacts. Additionally, although our unsupervised optimization framework effectively handles significant intra-class variations, semantic alignment becomes highly ambiguous when dealing with exceptionally disparate source-target geometries (e.g., morphing a chair into a bird). Future work will explore dynamic topology adaptation and connectivity-free guidance to further decouple the Gaussians from the static mesh scaffold to achieve unconstrained, cross-category 3D shape transformations

\noindent\textbf{Conclusion.} 
We introduce \textbf{SemMorph3D}, a novel semantic-aware framework that unifies 3D shape and texture morphing from multi-view images, bridging the gap between unstructured 3D Gaussian Splatting and structured mesh-based morphing. Our key innovation is a \emph{mesh-guided Gaussian morphing} strategy that anchors 3D Gaussians to semantic mesh patches. This approach bypasses the need for pre-aligned 3D assets and ensures that geometry and appearance are interpolated in a structurally consistent and texturally coherent manner. Through unsupervised learning guided by mesh topology, our method achieves state-of-the-art performance, outperforming existing 2D and 3D techniques in structural similarity, color consistency, and edge continuity. Furthermore, our framework exhibits strong intra-category generalization, enabling efficient morphing of novel object pairs without per-pair retraining. By generating visually faithful transformations, SemMorph3D sets a new standard for 3D morphing, opening exciting possibilities for visual effects and digital content creation.